# Catch Me If You Can: Identifying Fraudulent Physician Reviews with Large Language Models Using Generative Pre-Trained Transformers

Aishwarya Deep Shukla
Beedie School of Business,
Simon Fraser University

Laksh Agarwal
Faculty of Applied Science,
Simon Fraser University

Jie Mein (JM) Goh
Beedie School of Business,
Simon Fraser University

Guodong (Gordon) Gao
Carey Business School,
John Hopkins University

Ritu Agarwal
Carey Business School,
John Hopkins University

**Catch Me If You Can: Identifying Fraudulent Physician Reviews with Large Language Models Using Generative Pre-Trained Transformers**

## Abstract

The proliferation of fake reviews of doctors has potentially detrimental consequences for patient well-being and has prompted concern among consumer protection groups and regulatory bodies. Yet despite significant advancements in the fields of machine learning (ML) and natural language processing (NLP), there remains limited comprehension of the characteristics differentiating fraudulent from authentic reviews. This study utilizes a novel pre-labeled dataset of 38,048 physician reviews to establish the effectiveness of large language models in classifying reviews. Specifically, we compare the performance of traditional ML models, such as logistic regression and support vector machines, to generative pre-trained transformer models (GPT-3). Furthermore, we use GPT-4, the newest model in the GPT family, to uncover the key dimensions along which fake and genuine physician reviews differ.

Our findings reveal significantly superior performance of GPT-3 over traditional ML models in this context. Additionally, our analysis suggests that GPT-3 requires a smaller training sample than traditional models, suggesting its appropriateness for tasks with scarce training data. Moreover, the superiority of GPT-3's performance increases in the cold start context (i.e., when there are no prior reviews of a doctor). Finally, we employ GPT-4 to reveal the crucial dimensions that distinguish fake physician reviews. In sharp contrast to previous findings in the literature that were obtained using simulated data, our findings from a real-world dataset show that fake reviews are generally more clinically detailed, more reserved in sentiment, and have better structure and grammar than authentic ones.

## 1 Introduction

Online reviews are becoming increasingly prevalent and are exerting a significant influence on the consumer purchase process and final purchase decision [1–16]. Industry reports show that 78% of consumer look for reviews before making online purchases, and online reviews are now as much as seven times more powerful than advertisements in influencing their decisions [17]. Reviews influence decision making at different stages of purchase [18,19] and can even alter shoppers' preferences with regard to product attributes [20,21], with negative reviews having an extremely strong effect [22]. The rising economic influence of these reviews creates strong incentives for businesses to post fraudulent reviews to promote themselves and discredit their competitors [23]. Self-promotion in online reviews was first detected as early as 2004, when Amazon unmasked considerable self-reviewing by book authors [24]. Further, in 2011, it was reported that businesses were hiring workers on Amazon Mechanical Turk to post fraudulent five-star Yelp reviews on their behalf [25]. The issue of fraudulent reviews has not only attracted the attention of researchers [26] and the affected platforms, but regulatory authorities have also recognized the gravity of the problem and initiated punitive actions. For example, in 2013, the Attorney General of New York State fined companies $350,000 for faking their online reviews [27], and in 2021, the Federal Trade Commission (FTC) initiated proceedings against more than 700 businesses accused of participating in fraudulent review practices or concealing unfavorable feedback [28].

A major reason why fraudulent reviews are now such a serious concern is that they have affected the credibility of many review platforms by eroding consumer trust [26,29,30]. This problem is especially salient in the healthcare context, where patients make complex healthcare decisions based on the information obtained from online reviews. Given the potentially severe and adverse consequences for individuals' health outcomes, fake reviews in healthcare are particularly risky. Therefore, it is imperative to advance the methods of identifying fraudulent reviews. Existing studies of fraudulent review detection can be classified into behavioral and linguistic approaches (summarized in Table 1), both of which have unique strengths and weaknesses. Behavioral approaches focus on finding distributional anomalies of ratings and observing unusually correlated patterns in the review metadata. Such patterns could reveal, for example, collusion among a group of reviewers by checking whether these reviewers are systematically writing positive reviews for a group of products and/or negative reviews for a group of competing products [31,32]. As another example, behavioral approaches can be used to check whether a product has an abnormally high number of reviews from first-timers (reviewers who have never posted a review before) and singletons (reviewers who have posted just one review) [33]; all such reviews can then be flagged as suspicious.

To date, behavioral approaches are widely used in industry practice and have demonstrated higher accuracy than linguistic approaches [34]. However, the use of behavioral approaches specifically for the detection of fraud has a fundamental vulnerability. By definition, behavioral approaches attempt to decipher fraudulent behavior by converging on a pattern of behavior that could be considered an outlier. Skilled fraudsters understand that such deviations from normal behavior make them stand out as suspicious, and over time, they learn to closely emulate the behavior of genuine reviewers to ensure that their reviews are not filtered out, thereby gaming the system.

Alternatively, linguistic approaches detect fraudulent reviews by focusing on the differences in language patterns between genuine and fraudulent reviews. Research has shown that humans describing real events unconsciously use different language patterns than they do when describing imaginary events. For example, studies have shown that when writing fake reviews, people tend to use past tense and superlatives more frequently [35]. Machines can recognize such unconscious linguistic patterns and detect deception automatically in a wide range of contexts [36]. Since these linguistic approaches identify fraudulent reviews by looking for linguistic traits that fraudsters employ unknowingly, they are more robust against gaming.

However, the linguistic traits that distinguish a fraudulent review from an authentic one can be very subtle. To demonstrate the subtlety of these differences, we encourage readers to attempt classifying a selection of reviews from our dataset as either fake or genuine.[1] A sample of 10 reviews can be found in Table 2.

Although word-dictionary-based linguistic approaches appear promising when tested with simulated datasets [35], their efficacy in real-world data remains limited [37]. To address this challenge, our study adopts a state-of-the-art approach in using generative pre-trained transformer models (GPT-3) to consider the entire word sequence in the review text instead of focusing on a set of linguistic features like specific words. This approach not only extends the current literature on machine learning and neural network models but also reinforces the promising outcomes these tools have already demonstrated in natural language processing tasks, particularly in learning semantic representations [38,39]. We hypothesize that GPT-3 can capture the subtle semantic information embedded in review text, which is difficult to extract using existing methods.

[1] We also ran a simple lab experiment on a crowdsourced platform (AMT) asking people to rate a random sample of 20 reviews drawn from a larger set of 2,000 reviews. The results suggested that humans were more likely to label genuine reviews as fake and fake reviews as genuine than to label the classes correctly.

At the beginning of our research process, however, GPT-3's detection capabilities were far from certain: linguistic traits in fake reviews can be so subtle that it was unclear whether a generative model repurposed as a classifier would be powerful enough to detect them. We tested our novel approach on the dataset of doctor reviews obtained from one of the largest doctor platforms based in India, and we found that GPT-3 performs significantly better than traditional NLP and ML algorithms. Traditional machine learning algorithms exhibit lower accuracy in detecting fraudulent reviews, and their performance, as measured by F1 and F2 metrics, consistently falls short of GPT-3's superior results.

Furthermore, we tested the performance of all models on reduced training sample sizes, from only 100 reviews all the way up to 10,000 reviews, keeping the test data size constant. Our motivation for testing the performance of models on reduced sample sizes is grounded in the known difficulties firms face in acquiring datasets for a challenging task such as fake reviews of doctors, where the ground truth is hard to establish. For one, most companies do not have the considerable resources required to gather, organize, and annotate data. In addition, labeling a dataset with sufficient accuracy and consistency can be a time-consuming and expensive process. As a result, companies often struggle to obtain sufficiently large, labeled datasets, which in turn limits the effectiveness of machine learning algorithms trained on these datasets to smaller samples. We found that the performance gap between GPT-3 and other models widens when the sample size is very small, suggesting that GPT-3 is very useful when working with problems that have limited labeled data. The ability to achieve high performance with small datasets represents a crucial step towards democratizing machine learning and enabling more companies to benefit from its applications at a modest cost.

In addition, to investigate whether GPT-3 captures more profound distinctions between fake and genuine reviews than existing approaches, we retrained all our models with the testing dataset, which comprises reviews written for different doctors (in distinct clinics) by unique reviewers not included in the training dataset. In the literature [40–42], this scenario is referred to as the cold start problem – a highly relevant issue in the context of fake reviews, as it is crucial for platforms to effectively detect fraudulent reviews from first-time reviewers and for doctors who have not yet received any reviews. The observed large performance gap between GPT-3 and traditional machine learning models in this setting indicates that GPT-3 is able to utilize more in-depth differences between counterfeit and authentic reviews, transcending the performance of commonly used approaches based on the individual writing styles of reviewers or particular characteristics of specific doctors. As the final stage in our study, we deployed GPT-4's cutting-edge capabilities to discern the dimensions that help distinguish fake and genuine reviews (fake and genuine), a task that we attempted with topic modeling (LDA) but that resulted in very noisy and indiscernible differences.

Our study makes several contributions to related literatures on fake review detection and machine learning. It is the first to deploy a GPT-3 based classifier for online reviews, evaluate its performance with traditional machine learning algorithms, and use GPT-4 to extract the characteristics that differentiate fake reviews differ from genuine reviews. Furthermore, our study is unique in that unlike studies that utilize simulated data [35], we utilize labeled real-world data that allows us to uncover previously unrecognized differences in content between fake and genuine reviews. Specifically, we find that fake reviews tend to be more detailed, more reserved in sentiment, more clinically focused, and have better structure and grammar than authentic ones. This finding is in sharp contrast to earlier studies using simulated data, and it calls into question the general perception of the language of fake reviews as being vague, poorly structured, and being overly positive. The collection of our real-world data was made possible by authentication loopholes that allowed medical practices' staff to log in as patients and post fake reviews of their practice's doctors. Our labeling technique is completely exogenous to the techniques used in the classification model since the classification model only relies on text data and no metadata, ensuring that the estimated accuracy will be close to real-world accuracy.

Our study is situated in the context of a service of significant social consequence: healthcare providers. Since there are very few mechanisms that signal a doctor's quality to the public, online reviews have become a popular quality signal [43,44]. Such reviews are the digital equivalent of traditional word of mouth, which plays an important role in decision making, as highlighted in several studies [45–54]. A recent survey shows that almost three-quarters (72%) of patients have reported using online reviews as the first step in choosing their new doctor [55]. Since healthcare choices are high-stakes decisions, fake reviews that mislead patients pose a threat to patient safety. As a result, multiple news reports from outlets, such as the *New York Times*, the *Telegraph*, NBC, and the *Guardian*, have called for public attention to the issue of fake doctor reviews (Table 3). In light of this increasing concern, our study has significant implications for patient welfare.

# 2 Background

## 2.1 The rise of fraudulent reviews

Online word of mouth (WOM) is essential for businesses in many ways. First, as a quality signal, it is highly influential in driving sales and attracting new customers [56–61] despite the fact that it may be biased [7]. Second, studies have shown that positive online WOM motivates customer loyalty and drives community engagement [62]. Third, negative online WOM has an exceptionally detrimental impact on a firm's cash flows, profitability, and long-term stock prices [63]. Therefore, given the influence of online WOM on a firm's well-being, it is imperative that they maintain positive online ratings and reviews (types of online WOM).

However, striving for a consistent flow of positive ratings and reviews can be very costly as it requires firms to consistently meet or even exceed customer expectations. Unscrupulous businesses instead try to portray their products and services in a positive light by implanting false narratives (online reviews) of customer experiences of those products and services. Typically, such firms either ask their employees to post fraudulent online reviews or engage third-party reputation management companies to do so. Fraudulent reviews are quite widespread: recent data suggests that about 16% of reviews posted on large online review platforms such as Yelp are fake [23]. The production of fake reviews is a practice not limited to small firms, and large, reputable companies have also been implicated. For example, the Taiwan Fair Trade Commission caught Samsung hiring bloggers and students to write reviews that falsely praised its products and criticized its competitors' products [27].

Not surprisingly, fake reviews have become a major concern for consumer protection groups and regulatory authorities, as consumers can be misled into purchasing goods and services of questionable quality. Furthermore, as previously mentioned, when brought to light, fake reviews erode customer trust in the platforms that host them. Studies indicate that about 57% of customers are suspicious of online sellers that only have positive reviews and about 49% believe that sellers use unfair means (e.g., monetary incentives) to obtain these reviews; from this perspective, customers may consider platforms complicit for allowing such practices [64].

### 2.2 The cold start problem for reviews

The cold start problem refers to the difficulty of detecting fake reviews for products or services that have no prior reviews and to the difficulty of identifying true reviews posted by new users or reviewers without any prior history. There are only a handful of studies that examine the cold start review problem [41,42]. The goal of these papers is to develop algorithms to detect fake reviews in the context of the cold start review problem and to do so as soon as the reviews are posted. This is a significant challenge because fake review detection algorithms typically focus on features that are collected over a period of time. Without any prior history, machine learning algorithms underperform in detecting fake reviews in general and cold start reviews in particular. To address this ongoing challenge, more recent studies have examined the use of graph-based features to improve the accuracy of algorithms and ameliorate this problem in online reviews [41,42]. To overcome the limitations of the cold start fake review problem, we sidestep the need for historical feature and new feature data by examining extant linguistic features using a state-of-the-art algorithm, GPT-3.

### 2.3 Fraudulent reviews of doctors

Our study focuses on reviews that pertain to healthcare providers, a context where the detection of fraudulent reviews is complex due to the idiosyncrasies of the healthcare market. First, most patients consult their healthcare providers (doctors) infrequently: statistics suggest that an average U.S. citizen makes about four visits to a doctor each year [65]. This low frequency means that patients do not have much engagement with doctor review platforms, so there are few ratings and reviews for each doctor as compared to what might be typical with, for example, restaurants. To illustrate, on RateMDs.com, a popular doctor rating website, the average number of ratings per doctor is 3.2, and about half of the doctors have only one rating [43]. In contrast, on Yelp, the average restaurant has well over 40 ratings. Therefore, the number of fraudulent ratings/reviews required to bias consumer perception of a doctor is small – sometimes as low as one. Also, since patients only visit doctor review platforms infrequently, it is challenging for researchers to use browsing behavior to detect fraudulent reviewers.

Second, in contrast to consumer goods, healthcare can be classified as a credence good. This means that the consumers (patients) are less knowledgeable about the extent of the service needed (e.g., medication) or about the true competence of the provider. The provider acts as the expert in determining the requirements of the customer, and the customer can only judge healthcare services by the experience component of the service and not the credence component [66]. For example, patients tend to assess a surgeon based on observable bedside manner rather than on unobservable surgical skills. Since patients lack certainty about the quality of the service provided to them, they are less likely to publicly share their opinions. Also, because a review often reflects the patient's interactions with the doctor, there is a non-trivial chance that the doctor might be able to identify the reviewing patient even if the review is anonymous. These factors further discourage patients from reviewing their healthcare providers, reducing the total number of reviews obtained by doctors and making it more difficult to establish the authenticity of the reviews.

Third, from the perspective of a provider, the healthcare market offers limited means to signal one's quality [67–69]. For example, the doctors whose reviews were examined in this paper cannot advertise their service offerings or promote their quality like other businesses; this would be a violation of the Medical Council of India's code of medical ethics [70]. In addition, although multiple channels exist for patients to find information about a doctor's quality (such as Physician Compare, Checkbook or Best Doctor lists in the U.S.), WOM information remains the primary channel [55]. Due to the low volume of the WOM contributed by patients, building a real online reputation takes a long time. Therefore, there is a strong incentive for doctors to augment their online reputation artificially.

Finally, in the healthcare industry, the ticket size (the average revenue per transaction) is very high. In the U.S., surgeons earn thousands of dollars from a single patient. The higher economic value of a customer in healthcare, as compared with other highly reviewed service businesses such as restaurants and hotels, further intensifies providers' incentive to create fraudulent reviews, which has the knock-on effect of increasing the sophistication of the agents employed to create such fraudulent reviews.

### 2.4 Detecting fraudulent reviews using behavioral characteristics

Algorithms that rely on behavioral characteristics examine the deviation from the norms of observable characteristics to detect deception. Some of the most prevalent methods include tracking abnormal review frequency based on reviewer user ID, browser cookie ID and IP address, deviation in ratings, anomalies in geolocation, suspicious votes that mark the reviews as helpful, and the occurrence of recently received negative reviews [71–75]. All these behavioral features represent deviations from the expected behavior of a genuine reviewer.

Such behavioral approaches face challenges in detecting fake doctor reviews. First, patient privacy concerns sharply reduce the number of options available to platforms to verify a reviewer's authenticity. For example, some patients may not be comfortable using any single sign-on (SSO) service, such as Google, Facebook, or Microsoft, to sign into patient portals and doctor review sites. Second, patient privacy protection laws prohibit tracking of patients across the internet and forbid platforms that host reviews from requesting documentation, such as medical records, to verify the claims that patients made in their reviews. For instance, in the U.S., HIPAA privacy rules protect the privacy of individually identifiable health information, called protected health information (PHI) [76]. Third, the sparsity of doctor visits per patient makes it difficult to detect abnormalities in review posting behavior. Finally, due to the high rewards of establishing a good online reputation, skilled fraudsters can put in extra effort to closely emulate the observable behavior of genuine reviewers. For example, a highly-paid, resourceful reputation management firm engaged by a doctor can easily spoof an IP address or geolocation. Employing such a firm becomes more feasible when the volume of fraudulent reviews required to sway consumer opinion is low, as it is in healthcare. Unfortunately, the existing methods of fraud detection based on algorithms that utilize behavioral characteristics are limited by the multiple factors discussed here, so they are not resilient and may be easily gamed. This dilemma highlights the importance of advancing methods that utilize linguistic characteristics.

### 2.5 Detecting fraudulent reviews using linguistic characteristics

Fraud detection algorithms that rely on linguistic characteristics use the text of reviews to detect deception. Jindal and Liu [72], for example, discovered fraudulent reviews by identifying similarities among reviews using bigrams and trigrams (groups of two and three words). While the approach of using content similarities as clues to deception is potentially useful, it must be noted that fraudsters also understand that plagiarizing the content of other reviews would increase their odds of being detected by review filters, especially in markets where the volume of reviews is low. Therefore, in such markets where the incentives for establishing a favorable online reputation are high, fraudsters are likely to generate new content instead of plagiarizing existing content, making it harder to catch them.

Theoretically, linguistic traits hold substantial promise and can be resilient to gaming. Johnson and Raye [77], in their work on memory and perception, noted that truthful opinions are more likely to include sensory information and details about spatial configurations. Vrij et al. [78] also suggest that fraudsters have considerable difficulty in including spatial details. These approaches that examine linguistic traits do have promise, as they focus on finding language patterns that can explicitly indicate deception. For instance, Ott et al. [35] used a psycholinguistic deception detection method based on a linguistic inquiry and word count (LIWC) toolkit to derive features (words) in the reviews that can predict fraudulent behavior. Their work synthesized fake reviews: they paid freelancers to write about 400 imaginary reviews and collated real ones by scraping a popular travel site. While their study reported promising accuracy in classifying authentic and fake reviews using synthesized data, the replication of their approach with real-world data has highlighted its limitations [37], which are likely due to the systematic differences between synthetic and real-world data and to the study's small sample size.

In summary, despite the potential promise of linguistic approaches, existing studies have not demonstrated high accuracy with large real-world datasets. Our study aims to fill this knowledge gap by applying GPT-3 to find the subtle linguistic traits embedded in the review text to classify them and using GPT-4 to derive those traits.

### 2.6 Establishing ground truth

One major challenge in existing studies lies in establishing the ground truth, i.e., knowing with certainty which reviews are "fakes" and which are genuine. Previous studies on fraudulent review detection have mainly employed three methods for labeling data: using human evaluators, using a pre-labeled dataset that utilizes behavioral filters, or hiring individuals to fabricate reviews to create a repository of fraudulent reviews.

The first and most common method of data labeling involves human evaluation. For example, Xie et al. [79] employed human evaluators to read reviews from 53 stores and make decisions on the suspiciousness of the reviews. They then benchmarked their detection algorithm against human evaluators. Along similar lines, Mukherjee et al. [32] employed human judges to label groups of deceptive reviewers. The principal limitation in using human labeling as a proxy for distinguishing between fraudulent and genuine reviews is the low accuracy of human evaluation [79].

The second most common method of data labeling is to use available datasets of labeled reviews from leading review platforms such as Yelp [37]. However, the underlying filtering techniques used by such websites to label data remain imperfect [23,80]. Fundamentally, accuracy of the pre-labeled datasets is determined by the algorithms or heuristics, and the ground truth remains unknown, making it impossible to ascertain the accuracy of the labels.

The third method used to obtain labeled datasets is to scrape genuine published reviews from major online review portals and employ freelancers to write fraudulent reviews. Ott et al. [35], for example, acquired labeled genuine reviews by scraping a travel website, assuming all published reviews on that site to be genuine. They then obtained fake reviews by instructing freelancers (turkers) from Amazon Mechanical Turk (AMT) to pretend to be hotel employees whose managers asked them to make up customer reviews. In this manner, the study created a labeled dataset of genuine reviews (obtained from the website) and fraudulent reviews (written by freelancers). Although the researchers instructed the turkers to sound as realistic as possible in their pretense, the reviews generated by the turkers are dissimilar to actual fraudulent reviews posted by a sophisticated professional fraudster [37]. These differences could be due to the differences in writing skills, levels of economic affluence, motivations, and psychological states between the hired freelancers and the real fraudsters, making the latter better at faking customer reviews.

In reviewing the methods of data labeling, we also identify one common limitation in the existing work on linguistic approaches: the labeling method overlaps with the detection method, leading to an overestimation of performance. This confounding of the labeling and detection methods is evident in at least three ways. First, studies that use human evaluators to label data employ detection algorithms that are similar in principle to the approach used by human evaluators to label the reviews. For example, Lim et al. [33] labeled their data by showing the individual reviews, along with all reviews provided by each reviewer, to human evaluators, then asking them to judge the authenticity of the reviews. When human evaluators judge reviews to find fakes, they look for commonalities across reviews by the same reviewer. They also look for traits such as repeated words and phrases and other signs of deception. The detection algorithm used in the Lim et al. [33] study takes a similar approach by utilizing bigrams and trigrams to detect fraudulent reviews. This apparent similarity in the detection and labeling method makes it difficult to judge the accuracy of the detection algorithm as, in essence, it is an automation of a manual process. As another example, Mukherjee et al. [32] provided human evaluators with review content and asked them to label the reviews as fraudulent or genuine. Their detection algorithm uses features such as group content similarity to detect fraud, which is very similar to the method used by human judges to distinguish between genuine and fraudulent reviews.

Second, studies that use pre-labeled datasets from leading websites also suffer from a lack of independence of the labeling method from the detection method. To illustrate, in the Yelp dataset used by researchers, signs of deceptive language provide one basis for distinguishing between fraudulent and genuine reviews [37]. Studies such as those by Feng et al. [81] and Mukherjee et al. [37] use syntactic style information (i.e., grammatical similarity) in their fraudulent review detection algorithm. The data labeling method involving human judges also implicitly relies on the grammar of the reviews as a signal of fraudulent behavior. Therefore, the detection method is not independent of the labeling method, possibly overestimating model performance. Similarly, Li [82] used labeled data from a leading restaurant ratings website in China. The detection algorithm in the study examines metadata patterns (such as the geolocation IP addresses) to detect fraudulent reviews. This detection algorithm overlaps with the method used by the portal to label the data and is therefore not independent.

Finally, studies that label data by employing humans to create fraudulent reviews also suffer from the lack of independence of the data labeling method from the fraud detection method because of fundamental differences between the users who are paid to create the fraudulent reviews and the genuine reviewers. As previously mentioned, Ott [35] used AMT freelancers to write fraudulent reviews. The vocabulary used by freelancers likely differed not only from the vocabulary used by sophisticated fraudsters, but also from the vocabulary used by genuine reviewers. Here again, this would be due to fundamental differences in the abilities, level of economic affluence, motivations, and states of mind of the writers. Furthermore, since freelancers are asked to write favorable reviews, their reviews are likely to contain a high number of superlatives. The use of superlatives is, however, a factor utilized by the authors in differentiating genuine from fake reviews. Hence, the labeling and detection methods are not entirely independent, leading to overestimated performance.

To overcome the above limitations, we establish ground truth (detailed in Section 3.1) in a way that is independent of the detection algorithm. By doing so, we avoid infusing the knowledge of the labeling process into our algorithm, enabling a more objective and rigorous evaluation of its performance in detecting fraudulent reviews.

## 3 Data

Our data source is a leading online doctor search platform based in India. One of the authors of this study spent substantial time with the development team of this platform to adequately understand the system architecture. The platform has listings for over one-third of all private doctors in India's major cities. It allows patients to search and book appointments with doctors, and it subsequently requests that they provide feedback (a yes/no recommendation followed by a text review) about their consultation. Furthermore, it also accepts feedback from patients who have not booked an appointment through the portal.

Listed on the platform are doctors located in the major cities of India across different specialties. To write a review, one must register on the platform using a cell phone number (mandatory) and an email address (optional). In order to deter spamming, the platform enforces a restriction that prohibits patients from submitting multiple reviews for any doctor within the same city and specialty within a single month.

### 3.1 Labeling of reviews

For the duration of approximately one year, the platform experienced unintentional security vulnerabilities that enabled doctors (or their staff) to circumvent the authentication system and post reviews of themselves or the practice's doctors on behalf of their patients without the patient's knowledge. These vulnerabilities went undetected, creating a convenient mechanism for posting fraudulent reviews, which numerous doctors took advantage of. However, it also established a foolproof natural honeypot. We pinpointed the security flaws and informed the platform, and with their support, we undertook a comprehensive investigation of all reviews utilizing the available digital forensics data, including the persistent cookie ID, login credentials, session keys, and more. This facilitated the accurate identification of all published reviews that had taken advantage of the security vulnerabilities.

We marked the reviews that exploited these security gaps as fraudulent and the remaining as genuine. When these security vulnerabilities were present, they enabled a fraudster to bypass the validation process with relative ease, giving them minimal incentive to use more costly methods of fraud (such as establishing a new identity by obtaining a new mobile connection). The presence of this unintentional yet effective honeypot for an extended period offered a unique opportunity to generate a high-quality labeled dataset of both fraudulent and genuine reviews.

### 3.2 Data construction

Initially, we executed standard text sanitization processes, including trimming extra spaces and eliminating "junk" reviews, which were primarily composed of special characters. Furthermore, in adherence with the platform's minimum length policy for published reviews, we discarded reviews that were shorter than 50 characters. Duplicate reviews were also excluded, as fraudulent entities may have replicated existing reviews.

Subsequent to the text sanitization process, our dataset comprised 38,048 reviews, with 29,630 (77.88%) labeled as authentic and 8,418 (22.12%) as fake. Utilizing this primary dataset, we generated two datasets to examine the performance of different models. The first one, the standard sample (SS), incorporated reviews randomly chosen from the entire pre-processed dataset. The second one, the cold start sample (CS), contained only cold start reviews. Using this cold start criterion, we split the dataset into training and testing sets, ensuring no common doctors, reviewers, or clinics were shared between the two sets. This provided a close approximation to a true cold start dataset. With respect to the training sample size, we selected six discrete values – 100, 500, 1,000, 2,000, 5,000, and 10,000 – to enable a comparative analysis of the performance of multiple models across different sample sizes. In order to assess the models trained on the aforementioned sizes, we utilized a consistent test sample comprising 2,500 reviews.

## 4 Methods: Generative pre-trained transformer (GPT)

Our approach to detecting fraudulent reviews makes use of only the review text and does not use any of the metadata that captures the reviewer's posting behavior. This ensures that our method is less prone to being gamed if it is deployed in the real world. To make predictions, our method utilizes major techniques typically deployed in traditional machine learning algorithms and GPT-3. We aim to examine the potential of GPT-3, in comparison to traditional machine learning methods, to detect fake reviews by utilizing the subtle differences between the writing styles of fraudulent and genuine reviewers, as it is harder to game writing style than posting behavior.

GPT is a cutting-edge large language model (LLM) that has gained significant attention with the release of ChatGPT in Nov 2022. Created by OpenAI, it is capable of generating coherent and contextually relevant natural language text based on given input.

GPT is pre-trained on vast amounts of text data, with models ranging from 125 million to 175 billion parameters. Its primary goal is to predict the next word in a sentence by leveraging the statistical properties and relationships of words. GPT demonstrates exceptional performance across various natural language processing tasks, such as language translation, text summarization, and even question-answering. At the time of writing, there are four baseline GPT-3 models available for fine-tuning: ada, babbage, curie, and davinci, each with different capabilities. While the details of GPT-3 and newer models have not been published by OpenAI, we describe the technicalities of this class of GPT models referencing earlier versions of GPT (GPT-2).

In terms of its technical architecture, GPT is a large neural network consisting of multiple transformer layers, which are designed for parallel processing, resulting in faster training than traditional models like recurrent neural networks (RNNs). Developing a GPT model involves two steps: generative pre-training and discriminative fine-tuning. The generative pre-training step is an unsupervised learning process that utilizes a vast corpus of unlabeled data, establishing language representations by predicting the next word in a sequence [83]. During pre-training, GPT utilizes word embeddings as the language representation, leading to improved performance compared to static vectors like Word2Vec. Word embeddings are dense vector representations that allow for better computability among words and other factors. Traditional NLP techniques embed words, sentences, paragraphs, or documents as vectors based on word frequency, but this approach tends to ignore positional information. To address this limitation, GPT incorporates transformer architecture with self-attention mechanisms that capture and preserve positional information across input sequences.

In existing literature, several options exist for obtaining word embeddings. Researchers can use pre-trained word embeddings, such as GloVe [83] and fastText [84], or generate word embeddings using local text data [85]. In GPT, contextualized word embeddings, also known as dynamic embeddings, are characterized by anisotropic words that concentrate in a narrow cone in the embedding space. The upper layers are often more concentrated than the lower layers. These dynamic embeddings serve as input for the predictive model: word embeddings are fed into transformers that use a self-attention mechanism to assign weights based on the importance of each part of the input data.

Fine-tuning GPT with specific datasets can yield better performance, particularly when the pre-trained text corpus differs significantly from the target text data. During fine-tuning, a supervised learning process refines the model to perform specific tasks using labeled data by adding a task-specific output layer on top of the model trained to predict the target labels for the specific task. This discriminative fine-tuning process helps the model adapt to the nuances and context of the specific domain or task, allowing it to solve new tasks in the same domain [86,87].

The improvement in the performance of GPT-3 over GPT-2 is a result of multi-task pre-training, an extremely large dataset and extremely large models. Specifically, GPT-3 reportedly has 175 billion parameters and was trained on 45TB of data. A recent study suggests that GPT-3 can achieve reasonably good performance even without fine-tuning [88]. In our context, reviews of doctors written by Indian patients, presents unique challenges such as the common occurrence of medical jargon, systemic errors in grammar, and the inclusion of strong sentiments [86,87]. As such, large language models using GPT holds tremendous promise in tackling the fake review problem.

# 5 Results

## 5.1 Comparison against traditional machine learning

In our study, we assess the performance of the cutting-edge GPT-3 model relative to traditional machine learning methodologies[2]. GPT-3 is a large language model utilizing deep learning that has demonstrated promising results across a variety of natural language processing tasks, such as text summarization. Our objective is to evaluate the effectiveness of GPT-3 in comparison to widely used machine learning algorithms in text classification tasks including logistic regression, support vector machines (SVM), random forest and Extreme Gradient Boosting (XGBoost). Our evaluation is based on the precision, recall, F1-score, and F2-score metrics, which are frequently utilized to evaluate the performance of classification models [89].

[2] We utilize GPT-3 but not the GPT-4 models for this comparison because Open AI has not yet granted access for fine-tuning (training) GPT-4.

We choose these specific machine learning methods because they represent both linear and non-linear model types and exhibit high efficiency. Logistic regression, a prevalent baseline method, is routinely used as the initial model for machine learning tasks. Random forest is chosen as the second baseline model because of its widespread application and potential suitability for complex tasks such as fake review detection. Drawing from prior research in text classification[90,91], we incorporate support vector machines (SVM) and XGBoost as additional baseline models, given their established efficacy in text classification.

Table 4 and Figures 1-2 report the performance of various models on the standard sample. F1 and F2 metrics indicate that the performance of GPT-3 considerably surpasses that of other models, particularly at smaller sample sizes. The F1 score for GPT-3 is noted as 0.418, and the F2 score is ascertained as 0.383 with a training sample size of 100. Conversely, the next highest performing model, XGBoost, attains an F1 score of 0.280 and an F2 score of 0.226. Logistic regression, however, fails to detect any fake reviews, yielding F1 and F2 scores of 0.0.

As sample sizes increase incrementally, this performance trend persists, although the gap between GPT-3 and traditional models narrows. The average difference in F1 scores between traditional models and GPT-3 gradually decreases from 0.313 to 0.089 as the sample size expands from 100 to 10,000. Nevertheless, GPT-3 consistently outperforms all other models regardless of sample size.

These findings, specifically the F1 and F2 metrics, suggest superior performance of the GPT-3 model compared to the other models. It is essential to acknowledge, as highlighted in the introduction, that all algorithms employed in this study exclusively utilize linguistic features (i.e., review text) while deliberately excluding behavioral features. This deliberate exclusion aims to mitigate the susceptibility of our classification approach to manipulation by malicious actors.

### 5.2 Comparison against traditional machine learning for cold start problem

As discussed earlier, the cold start problem poses a significant challenge due to the potentially diminished usability of existing models when they are applied to data from new reviewers. We compare the performance of traditional machine learning and a GPT-3 based classifier on the cold start sample, with results presented in Table 5 and Figures 3-4. The sample of data for these models is set up such that all reviews from reviewers in test and training data are mutually exclusive, as are reviews of the same doctors and clinics. In other words, if the model has been trained with reviews written by a certain reviewer for a particular doctor working out of a certain clinic, the model is tested on reviews written by other reviewers for different doctors who are practicing in different clinics. This contrasts with the sample in Table 4, wherein the reviews are not common between the training and testing samples, but the reviewers, doctors, and clinics may be common.

In these results, as anticipated, we observe a decline in the performance of all models. However, the performance gap between GPT-3 and other models is more pronounced. For example, the F1 score for the GPT-3 model is 0.368, compared to the XGBoost's F1 score of 0.126 for a sample size of 100. Likewise, the F2 score for the GPT-3 model is 0.285, as opposed to 0.100 for XGBoost. It is worth noting that every other model tested (LR, RF, and SVM) failed to correctly predict even a single fake review, resulting in F1 and F2 scores of 0.0 for each. Although these models demonstrate improvement with increasing sample size, the trend of GPT-3 outperforming all other models persists. For instance, with a training set size of 2,000, GPT-3 achieves an F1 score of 0.467, while LR, RF, XG, and SVM score 0.121, 0.016, 0.328, and 0.248, respectively.

These findings suggest that GPT-3 can discern and recognize certain subtle characteristics prevalent across fake reviews in general, transcending the stylistic peculiarities of individual reviewers or doctors. Once again, these results underscore the superior performance of the GPT-3 model in comparison to the other models.

### 5.3 Discerning the differences in content between fake and genuine reviews

We outline the content differences between fake and genuine reviews in Table 6. To obtain these differences, we first attempted the traditional topic modeling method using latent Dirichlet allocation (LDA), obtaining results that were extremely noisy and hard to discern. We further tried GPT-3 and GPT-4, with the former producing the same outcome as LDA. On the other hand, GPT-4 considerably outperforms LDA and GPT-3 in generating useful insights in discerning these differences as presented in Table 6.

Prior literature has identified several linguistic traits that differentiate genuine reviews from fake ones. These differences can be traced to aspects such as specificity, sentiment polarity, readability, content focus, and language complexity [35,92,93]. According to these studies, genuine reviews generally demonstrate greater specificity and descriptiveness in their language, as they are based on actual experiences with the product or service in question. Researchers like Mukherjee et al. [34] and Ott et al. [35] have found that fake reviews often employ vague or general terms because they use a crowdsourcing approach to obtain a synthetic dataset where authors of the reviews lack firsthand knowledge of the subject matter. In contrast, we find from our real-world dataset of actual fake reviews that these fakes are more specific, more detailed and more comprehensive with regard to the medical condition of the patient, the diagnostic tests and the treatment plans. This suggests that unlike datasets based on simulated or synthetic fake reviews, real-world fake reviews demonstrate greater specificity.

In terms of sentiment polarity, the extant literature suggests that genuine reviews typically exhibit more balanced sentiment, addressing both the positive and negative aspects of a product or service. Jindal and Liu [72] and Li et al.[31] have noted that fake reviews tend to display extreme sentiment, either overly positive or negative, as their primary goal is often to artificially inflate ratings or tarnish a competitor's reputation. In contrast, we find that fake reviews exhibit more reserved and measured sentiment, possibly to increase their credibility.

Furthermore, studies suggest that genuine reviews tend to have a more natural writing style, with language patterns and syntax that are consistent with everyday communication. Researchers such as Mayzlin et al. [94] have observed that fake reviews may exhibit unusual language patterns, grammatical inconsistencies, or an excessive use of superlatives, thereby compromising their readability and authenticity. Our findings indicate that genuine reviews display a more conversational and casual language style, a finding consistent with previous literature. However, in contrast to prior research, we also observe that fake reviews exhibit better structure and fewer grammatical inconsistencies.

Lastly, in terms of content and topical focus, studies have found that genuine reviews emphasize product- or service-specific details while fake reviews focus on peripheral information [35,37]. In contrast, in our dataset we find that fake physician reviews provide detailed healthcare-specific issues and treatment, whereas genuine reviews tend to discuss peripheral attributes like clinic environment and the doctor's mannerisms.

## 6 Discussion and Conclusion

Our work is motivated by the economic significance of fraudulent reviews and the lack of robust methods for detecting them and understanding the underlying traits. This is a multifaceted and longstanding problem. First, up to this point the tools on hand have not proven altogether satisfactory in identifying fraudulent reviews, and they have been even less capable of identifying the characteristics that distinguish fake from genuine reviews. Second, we highlight the difficulty inherent in establishing the ground truth for detecting fraudulent reviews. Finally, we note endogeneity bias in the previous studies: the detection algorithm tends to be informed by the labeling strategy, which might lead to overestimated model performance and incorrect determination of distinguishing features.

The core contribution of our study is its use of large language models (GPT-3 and GPT-4) to successfully identify fraudulent reviews and their distinctive characteristics. We were able to leverage these models, in conjunction with a real-world dataset obtained via an existing security vulnerability which allowed us to successfully establish ground truth, to capture the specific linguistic traits in real-world fake reviews. We demonstrate that this approach significantly outperforms traditional machine learning methods in identifying fake reviews and in discerning the differences between real and fake. Our study advances the literature on using linguistic techniques to detect fraudulent reviews and provides evidence on the differences between the real-world fake and genuine reviews. These findings may contradict the current general understanding of the characteristics of fake reviews, which is largely based on synthetic data.

Our work has significant practical implications for online platforms: it provides empirical evidence of the efficacy of large language models in detecting fraudulent reviews and discerning the subtle distinctions between fakes and the real. Online platforms can capitalize on this knowledge to uphold the integrity and credibility of their review systems, thereby bolstering their reputation and enhancing user confidence. Moreover, since we derived substantial insights from a real-world dataset of counterfeit reviews that were inadvertently obtained through a temporary security vulnerability exploited by fraudsters, this suggests that platforms may strategically employ this method. In certain cases, authentication loopholes could be temporarily preserved to gather valuable training data on the unique linguistic features of fraudsters operating on their platform. Additionally, our findings highlight the importance of implementing safeguards within generative AI models to prevent the creation of fake reviews containing sensitive information, such as individuals' medical details. This would keep AI-generated fake reviews vague and generic and would further ensure the safety and reliability of any given platform's review system.

We acknowledge the limitations of our work. First, our data is limited to the period the security vulnerability existed. While our fraud detection model utilizes only the text of the data and has no input from the signals that were used to label the data, it is possible that after the security vulnerabilities were fixed, the fraudsters may have changed their writing style, and this would impact model performance. Second, our dataset contained fake reviews that were written by humans and detected by GPT-3. In the future, generative models may be used to correct the grammar of genuine reviews and to create fake reviews based on human-provided prompts. In such scenarios, where both authentic and fraudulent reviews can be processed or generated by similar generative models, detecting fake reviews might become more challenging. Third, GPT-3 has a training size constraint: the model can be fine-tuned with a maximum of 10 million characters. This restriction, when using three epochs, results in a maximum possible GPT-3 training sample size of slightly over 10,000 reviews, limiting our ability to test on the entire dataset. Finally, in our performance comparison of traditional machine learning algorithms and GPT-3, we omit neural network deep learning models such as CNN and RNN. These models often need substantial training data to achieve optimal performance, and obtaining very large, labeled datasets of fake reviews is very challenging if not impossible. As a result, the applicability of neural network models in this context might be constrained.

In conclusion, we note that the proliferation of fake content on online platforms can have severe adverse consequences for consumer welfare, and more research is needed to detect such content. Developing a method to detect such fraudulent behavior and pinpointing the key characteristics distinguishing fake reviews is the first step towards eliminating it, and generative models such as GPT-3 offer considerable promise of automatic detection with high accuracy. We believe, however, that unless it is combined with follow-up steps to penalize the doctors who fake their reviews and unless the penalty is harsh enough to act as a deterrent, a detection algorithm will not do much to truly fix the problem. This applies across a variety of settings, and it is especially relevant since methods for conducting fraud keep getting more sophisticated. The new insights from our study in regard to the differences between fake and genuine reviews could be potentially valuable in educating individuals on the distinctive characteristics of fraudulent reviews, thereby dispelling preconceived notions and enhancing classification accuracy. As previously mentioned, we conducted an experiment wherein human participants were asked to classify reviews; their accuracy was notably low, with a higher rate of misclassification in both genuine and fake reviews. This classification performance may significantly improve if individuals are informed about the specific characteristics of fraudulent reviews, such as those identified by the GPT-4 model. Future experiments can be designed to investigate and establish the potential improvement in classification accuracy after educating individuals on these characteristics. Finally, we note that more advanced generative pre-trained transformer models such as GPT-4 would further improve the accuracy of classification. As of the date of this writing, although the GPT-4 model is available for API access, it does not allow for fine-tuning (training) and therefore cannot be used on the classification task, although it can be utilized the discern the differences between the two sets of pre-labeled data. When fine-tuning of GPT-4 becomes available, the performance of GPT-4 can be compared to the performance of GPT-3 for identifying fake physician reviews.

**Tables and Figures**

**Table 1: Summary of Existing Literature on Review Fraud Detection**

| Study | Approach | Methodology | Dataset | Reported accuracy |
|---|---|---|---|---|
| [71] | Linguistic | Identify reviews that are duplicates, have extreme sentiment, and have high similarity between review and product description | 5.8 million reviews from Amazon | Accuracy measure is not proposed as ground truth is not established. |
| [87] | Linguistic | Identify reviews that are duplicates or are highly similar | 100,000 Amazon reviews | Roughly 10-15% of reviews have high similarity. The follow-up qualitative survey reveals that in many cases, reviewers consult existing reviews to draft their own review. |
| [33] | Behavioral | Identify reviewers who target certain products and product groups and whose ratings have significant deviation from others | 11,038 Amazon reviews | Absence of ground truth labels of users leads them to compare their method to labels generated by experts. They find that their models match experts' labels more closely than "report helpful" metrics. |
| [35] | Linguistic | SVM and Naive Bayes classifier | 400 synthesized fraudulent reviews from MTurk and 400 published reviews from TripAdvisor | They report an 89.8% accuracy with LIWC and SVM with bigrams. |
| [81] | Behavioral | Distributional anomalies of ratings to label fraudulent reviews | Reviews scraped from Amazon and TripAdvisor; they subsequently test their algorithm on the dataset of [35] | They report 74% accuracy comparing their algorithm on the Ott et al. [35] dataset. |
| [81] | Linguistic | Context-free grammar (CFG) parse trees | Same data as [35] | They report an accuracy of 91.2%. |
| [32] | Behavioral (at group level) | Detecting groups of fraudulent reviewers by observing behaviors of groups of individuals who provide similar ratings for the same groups of products over time | 109,518 reviews from Amazon; in addition to temporal signals, human judges are used to label data on a spam index | They report a 0.95 area under the curve (AUC). |
| [79] | Behavioral | Observe unusually correlated patterns in the time of review posting and in the content of reviews | 408,469 reviews from resellerratings.com | Their baseline is human evaluation of spam, and they report a 61.11% precision and 75.86% recall on suspicious reviews. Furthermore, they report that human evaluators have a very low concurrence rating, implying that human evaluators often disagree. |
| [95] | Behavioral | Markov random field model and loopy belief propagation algorithm | 210,761 reviews from Amazon | They report an overall accuracy of 71.2%. |
| [34] | Behavioral and linguistic | Unigrams and bigrams, review length, reviewer deviation, percentage of positive reviews of a product, etc. | Yelp reviews of 85 hotels and 130 restaurants in the Chicago area | They benchmark the accuracy of their algorithm against the labels on Yelp (filtered reviews and published reviews) and report 86.1% accuracy. |
| [96] | Behavioral and linguistic | User data – personal, social, activity, past review text; review length | Yelp reviews of consumer electronics businesses in New York, San Francisco, LA and Miami | They reached a maximum F-score of 82%. |
| [97] | Linguistic | Product word composition classifier (PWCC), $\text{TRIGRAMS}_{\text{SVM}}$ classifier, and $\text{BIGRAMS}_{\text{SVM}}$ classifier | Subset of Amazon.com reviews scraped in June 2006 | They reported f, p, r values of 0.772, 0.764, and 0.781 using the best method. |
| [98] | Behavioral and linguistic | Unigrams, bigrams, test statistics, sentiment evaluation, metadata features – rating, rating deviation, singleton, burst features (density, mean rating deviation, deviation from the local mean, early time frame), textual features, rating features, temporal features | Same datasets as [99] Yelp datasets of restaurants in NYC and businesses located in the U.S., including New York, New Jersey, Vermont, Connecticut and Pennsylvania | They achieved an overall accuracy of 80.6, with precision of 77.6, recall of 86.1, and f-score of 81.6. |

**Table 2: Examples of Fraudulent and Genuine Doctor Reviews**

| S No. | Review |
|---|---|
| 1 | Dr Priya Patel is conscientious and efficient. She takes pains to understand the exact needs of the patient and is not pushy or ‘over advisory’. Her pleasant disposition takes pain out of the drill! |
| 2 | Dr. Anjali spent considerable time in examining and understanding the subject. She was extremely good in creating a soothing feeling in the patient. I am glad I am acquainted with her. |
| 3 | Dr. Ravi is a very good doctor. He is very dedicated towards his profession. He’s never in hurry to see more and more patients, in fact gives response to all queries very patiently. |
| 4 | Excellent!! The doctor is very friendly hence makes patient comfortable sharing all problems and also she probes the patient to diagnose the cause in depth. Explains everything in details. |
| 5 | It has been a pleasant experience working with Doctor, his treatment are of great quality. His office is well organized and things are very methodically planned. |
| 6 | Dr. Rajesh comes across as a very friendly and experienced doctor. He has been regularly seeing my parents, spouse and kids and we are all very happy with our experience. Highly recommended. |
| 7 | It was wonderful... The way she listen me and gave the proper cure, was fabulous.. Her expertise in her domain is incredible.. I wana thank her for the cure... |
| 8 | My experience was very good. Very kind and supportive doctor. hey provide very familiar environment surely i will recommend him to my friends and family members. |
| 9 | Very friendly and listened to our problems. Patiently replied to all our queries. We trust him now and would recommend him. The clinic is also centrally located and it is easy for us. |
| 10 | Very good Dr, kind, soft, knowledgeable and experienced. Spent more time to find my problem, explained about my problem and treatment. Solved my spinal problems by electrotherapy and exercises. |

Note: The reviews are presented verbatim. However, the names have been changed to withhold doctor’s identity. Reviews 3, 5, 7, 8, and 10 are fake, while the rest are genuine.

**Table 3: News Articles on Fraudulent Doctor Reviews**

| Source | Summary |
|---|---|
| [100] | Due to a psychotherapist's difficulty in gathering testimonials from his patients to use online, he has resorted to writing a submitting a few ratings quoting his own patients on his behalf |
| [101] | Anti-vaccinators left fake, negative reviews of a pediatric clinic on Facebook, Yelp, and Google |
| [102] | General practitioners are writing fake reviews about themselves on the NHS Choices website |
| [103] | Doctors have reported experiencing both negative and positive reviews associated with themselves on doctor review websites |
| [104] | Dozens of medical clinics in the U.S. were found to have used a Bangladeshi review broker who recruited people to write fake positive reviews for them |
| [105] | "Celebrity dentist" with hundreds of five-star reviews had reviews written by review rings that engage in compensated activities |
| [106] | Family physician frustrated with fellow physicians after witnessing them using fake positive reviews for their practice and posting fake negative reviews for their competition |

**Table 4: Results from the Full Dataset**

| Model | Training Sample Size | Precision | Recall | Specificity | Sensitivity | F1 Score | F2 Score | Δ F1 GPT-3 vs Model |
|---|---|---|---|---|---|---|---|---|
| LR | 100 | 0.000 | 0.000 | 1.000 | 0.000 | 0.000 | 0.000 | 0.418 |
| RF | 100 | 0.875 | 0.012 | 0.999 | 0.012 | 0.024 | 0.015 | 0.394 |
| XG | 100 | 0.466 | 0.201 | 0.931 | 0.201 | 0.280 | 0.226 | 0.138 |
| SVM | 100 | 0.735 | 0.063 | 0.993 | 0.063 | 0.116 | 0.077 | 0.302 |
| GPT-3 | 100 | 0.493 | 0.363 | 0.889 | 0.363 | 0.418 | 0.383 | |
| LR | 500 | 0.844 | 0.047 | 0.997 | 0.047 | 0.089 | 0.058 | 0.470 |
| RF | 500 | 0.957 | 0.117 | 0.998 | 0.117 | 0.208 | 0.142 | 0.351 |
| XG | 500 | 0.574 | 0.326 | 0.928 | 0.326 | 0.416 | 0.357 | 0.143 |
| SVM | 500 | 0.792 | 0.305 | 0.976 | 0.305 | 0.441 | 0.348 | 0.118 |
| GPT-3 | 500 | 0.593 | 0.529 | 0.892 | 0.529 | 0.559 | 0.540 | |
| LR | 1000 | 0.894 | 0.162 | 0.994 | 0.162 | 0.275 | 0.194 | 0.284 |
| RF | 1000 | 0.971 | 0.234 | 0.998 | 0.234 | 0.377 | 0.276 | 0.182 |
| XG | 1000 | 0.657 | 0.361 | 0.944 | 0.361 | 0.466 | 0.397 | 0.093 |
| SVM | 1000 | 0.830 | 0.358 | 0.978 | 0.358 | 0.500 | 0.404 | 0.059 |
| GPT-3 | 1000 | 0.702 | 0.464 | 0.941 | 0.464 | 0.559 | 0.498 | |
| LR | 2000 | 0.829 | 0.279 | 0.983 | 0.279 | 0.418 | 0.322 | 0.224 |
| RF | 2000 | 0.964 | 0.326 | 0.996 | 0.326 | 0.488 | 0.376 | 0.154 |
| XG | 2000 | 0.705 | 0.435 | 0.946 | 0.435 | 0.538 | 0.471 | 0.104 |
| SVM | 2000 | 0.811 | 0.419 | 0.971 | 0.419 | 0.552 | 0.464 | 0.089 |
| GPT-3 | 2000 | 0.682 | 0.606 | 0.916 | 0.606 | 0.641 | 0.619 | |
| LR | 5000 | 0.844 | 0.386 | 0.979 | 0.386 | 0.529 | 0.433 | 0.157 |
| RF | 5000 | 0.957 | 0.424 | 0.994 | 0.424 | 0.588 | 0.477 | 0.099 |
| XG | 5000 | 0.838 | 0.478 | 0.972 | 0.478 | 0.609 | 0.523 | 0.077 |
| SVM | 5000 | 0.871 | 0.435 | 0.981 | 0.435 | 0.580 | 0.483 | 0.107 |
| GPT-3 | 5000 | 0.751 | 0.632 | 0.938 | 0.632 | 0.686 | 0.652 | |
| LR | 10000 | 0.863 | 0.428 | 0.980 | 0.428 | 0.572 | 0.476 | 0.141 |
| RF | 10000 | 0.984 | 0.536 | 0.997 | 0.536 | 0.694 | 0.589 | 0.019 |
| XG | 10000 | 0.895 | 0.489 | 0.983 | 0.489 | 0.632 | 0.537 | 0.081 |
| SVM | 10000 | 0.923 | 0.442 | 0.989 | 0.442 | 0.597 | 0.493 | 0.116 |
| GPT-3 | 10000 | 0.886 | 0.597 | 0.977 | 0.597 | 0.713 | 0.639 | |

Note: To evaluate the models trained on the mentioned training sample sizes, we utilized a uniform test set consisting of 2,500 reviews.

**Table 5: Results from the Cold-Start Dataset**

| Model | Training Sample Size | Precision | Recall | Specificity | Sensitivity | F1 Score | F2 Score | Δ F1 GPT-3 vs Model |
|---|---|---|---|---|---|---|---|---|
| LR | 100 | 0.000 | 0.000 | 1.000 | 0.000 | 0.000 | 0.000 | 0.368 |
| RF | 100 | 0.000 | 0.000 | 1.000 | 0.000 | 0.000 | 0.000 | 0.368 |
| XG | 100 | 0.219 | 0.088 | 0.922 | 0.088 | 0.126 | 0.100 | 0.243 |
| SVM | 100 | 0.000 | 0.000 | 1.000 | 0.000 | 0.000 | 0.000 | 0.368 |
| GPT-3 | 100 | 0.719 | 0.247 | 0.976 | 0.247 | 0.368 | 0.285 | |
| LR | 500 | 1.000 | 0.002 | 1.000 | 0.002 | 0.004 | 0.002 | 0.371 |
| RF | 500 | 0.500 | 0.002 | 1.000 | 0.002 | 0.004 | 0.002 | 0.371 |
| XG | 500 | 0.467 | 0.294 | 0.916 | 0.294 | 0.361 | 0.317 | 0.014 |
| SVM | 500 | 0.708 | 0.068 | 0.993 | 0.068 | 0.124 | 0.083 | 0.251 |
| GPT-3 | 500 | 0.623 | 0.268 | 0.960 | 0.268 | 0.375 | 0.302 | |
| LR | 1000 | 0.813 | 0.026 | 0.999 | 0.026 | 0.050 | 0.032 | 0.374 |
| RF | 1000 | 0.643 | 0.018 | 0.998 | 0.018 | 0.035 | 0.022 | 0.390 |
| XG | 1000 | 0.523 | 0.314 | 0.929 | 0.314 | 0.393 | 0.341 | 0.032 |
| SVM | 1000 | 0.714 | 0.140 | 0.986 | 0.140 | 0.234 | 0.167 | 0.190 |
| GPT-3 | 1000 | 0.535 | 0.352 | 0.924 | 0.352 | 0.425 | 0.378 | |
| LR | 2000 | 0.733 | 0.066 | 0.994 | 0.066 | 0.121 | 0.081 | 0.346 |
| RF | 2000 | 0.500 | 0.008 | 0.998 | 0.008 | 0.016 | 0.010 | 0.451 |
| XG | 2000 | 0.585 | 0.228 | 0.960 | 0.228 | 0.328 | 0.260 | 0.139 |
| SVM | 2000 | 0.673 | 0.152 | 0.982 | 0.152 | 0.248 | 0.180 | 0.219 |
| GPT-3 | 2000 | 0.644 | 0.366 | 0.950 | 0.366 | 0.467 | 0.401 | |
| LR | 5000 | 0.714 | 0.150 | 0.985 | 0.150 | 0.248 | 0.178 | 0.159 |
| RF | 5000 | 0.444 | 0.016 | 0.995 | 0.016 | 0.031 | 0.020 | 0.376 |
| XG | 5000 | 0.676 | 0.234 | 0.972 | 0.234 | 0.348 | 0.269 | 0.059 |
| SVM | 5000 | 0.684 | 0.104 | 0.988 | 0.104 | 0.181 | 0.125 | 0.226 |
| GPT-3 | 5000 | 0.646 | 0.297 | 0.959 | 0.297 | 0.407 | 0.333 | |
| LR | 5000 | 0.714 | 0.150 | 0.985 | 0.150 | 0.248 | 0.178 | 0.159 |
| RF | 5000 | 0.444 | 0.016 | 0.995 | 0.016 | 0.031 | 0.020 | 0.376 |
| XG | 5000 | 0.676 | 0.234 | 0.972 | 0.234 | 0.348 | 0.269 | 0.059 |
| SVM | 5000 | 0.684 | 0.104 | 0.988 | 0.104 | 0.181 | 0.125 | 0.226 |
| GPT-3 | 5000 | 0.646 | 0.297 | 0.959 | 0.297 | 0.407 | 0.333 | |
| LR | 10000 | 0.737 | 0.168 | 0.985 | 0.168 | 0.274 | 0.199 | 0.081 |
| RF | 10000 | 0.571 | 0.032 | 0.994 | 0.032 | 0.061 | 0.039 | 0.294 |
| XG | 10000 | 0.727 | 0.202 | 0.981 | 0.202 | 0.316 | 0.236 | 0.038 |
| SVM | 10000 | 0.702 | 0.080 | 0.992 | 0.080 | 0.144 | 0.097 | 0.211 |
| GPT-3 | 10000 | 0.597 | 0.252 | 0.958 | 0.252 | 0.354 | 0.285 | |

Note: To evaluate the models trained on the mentioned training sample sizes, we utilized a uniform test set consisting of 2,500 reviews.

**Table 6: Differences in Content Between Genuine and Fake Reviews, as Identified and Described by GPT-4**

| Dimension | Group 0 (Genuine) | Group 1 (Fake) |
|---|---|---|
| Topics | These reviews generally contain positive feedback, often including phrases like "very good," "fantastic," or "amazing." | The reviews in this group also contain positive feedback but tend to focus more on the doctor's knowledge, experience, and professionalism. |
| Topics | Patients mention satisfaction with their experience, such as feeling comfortable, receiving detailed explanations, or seeing positive results.<br><br>There is a strong focus on the doctor's approach and bedside manner, including being friendly, caring, patient, and understanding.<br><br>Some reviews mention modern equipment and a well-organized clinic environment. | Patients in this group describe the doctor as being good at providing explanations, guiding them through the treatment process, and giving unbiased advice.<br><br>There is more emphasis on the doctor's ability to diagnose, treat, and manage conditions effectively.<br><br>Some reviews mention the doctor's willingness to work cooperatively with other healthcare professionals and their ability to mentor others. |
| Sentiment | Both groups exhibit predominantly positive sentiment. Group 0 reviews tend to contain more enthusiastic language and exclamation marks, often expressing a higher level of excitement or satisfaction with their experiences. | Both groups exhibit predominantly positive sentiment. Group 1 reviews, while still positive, convey a more reserved and measured sentiment, focusing on the doctor's professional attributes and expertise. |
| Writing Style | Group 0 reviews are more likely to contain casual and conversational language. Patients often describe their personal experiences and feelings, which creates a more informal and relatable tone. | Group 1 reviews tend to use more formal language and focus on the doctor's professional qualities, such as their knowledge, experience, and ability to diagnose and treat conditions effectively. The writing in this group is more structured, with sentences that are often more detailed and comprehensive. |
| Sensitivity of Content | Group 0 reviews tend to be less sensitive in terms of the content shared. The focus is more on the patients' experiences and feelings about their interactions with the | Group 1 reviews contain more sensitive information, as they often discuss specific medical conditions, treatment plans, and healthcare-related decisions. Patients in this |
| | doctor, the clinic environment, and the overall treatment process. | group are more likely to share detailed information about their health issues and the doctors' recommendations. |
| Extent of Medical Information | Group 0 reviews typically reveal less medical information. They may mention the general reasons for the visit, such as a check-up or a specific issue, but they do not go into great detail about the diagnosis or treatment. | Group 1 reviews, on the other hand, often contain more detailed medical information. Patients in this group are more likely to discuss the specific conditions they sought treatment for, the diagnostic tests they underwent, and the treatment plans recommended by the doctor. This group also tends to mention the doctor's knowledge, experience, and ability to provide evidence-based advice more frequently. |
| Specificity of Treatment Outcomes | Group 0 reviews tend to focus more on the overall experience and satisfaction, with less emphasis on specific treatment outcomes. | Group 1 reviews provide more information about treatment outcomes, often discussing improvements in symptoms or the effectiveness of the doctor's recommendations. |
| Focus on Staff and Clinic Environment | Group 0 reviews mention staff and clinic environment more frequently, emphasizing the role of support staff and the overall experience at the clinic. | Group 1 reviews focus more on the doctor's professional qualities and expertise, with less emphasis on staff and clinic environment. |
| Follow-Up Care and Communication | There is no clear distinction between the two groups in terms of follow-up care and communication, as both groups have reviews mentioning the doctors' willingness to communicate and provide guidance outside of appointments. | There is no clear distinction between the two groups in terms of follow-up care and communication, as both groups have reviews mentioning the doctors' willingness to communicate and provide guidance outside of appointments. |
| Frequency of Visits | Group 0 reviews don't provide much information about the frequency of visits, but the focus on the overall experience suggests that they could be a mix of one-time appointments and ongoing care. | Group 1 reviews seem to be more related to ongoing care, as they often discuss specific conditions, diagnostic tests, and treatment plans in greater detail. |

- While both groups have positive sentiment, Group 0 reviews exhibit more enthusiasm and excitement, with a more casual and conversational writing style. In contrast, Group 1 reviews have a more reserved sentiment and a formal writing style that focuses on the doctors' professional attributes.
- Group 0 reviews are less sensitive in terms of content and reveal less medical information, focusing more on the patients' experiences and feelings. Group 1 reviews contain more sensitive content and detailed medical information, with an emphasis on the doctors' professional attributes and expertise in managing specific health conditions.
- Group 0 reviews tend to focus more on the overall experience, staff, and clinic environment, with less emphasis on specific treatment outcomes. Group 1 reviews, on the other hand, provide more information about treatment outcomes and the doctor's professional qualities, with less focus on staff and clinic environment. There is no clear distinction between the two groups in terms of review length, patient demographics, or follow-up care and communication. The frequency of visits might be higher for Group 1.

**Fig 1: F1 Scores from Full Dataset**

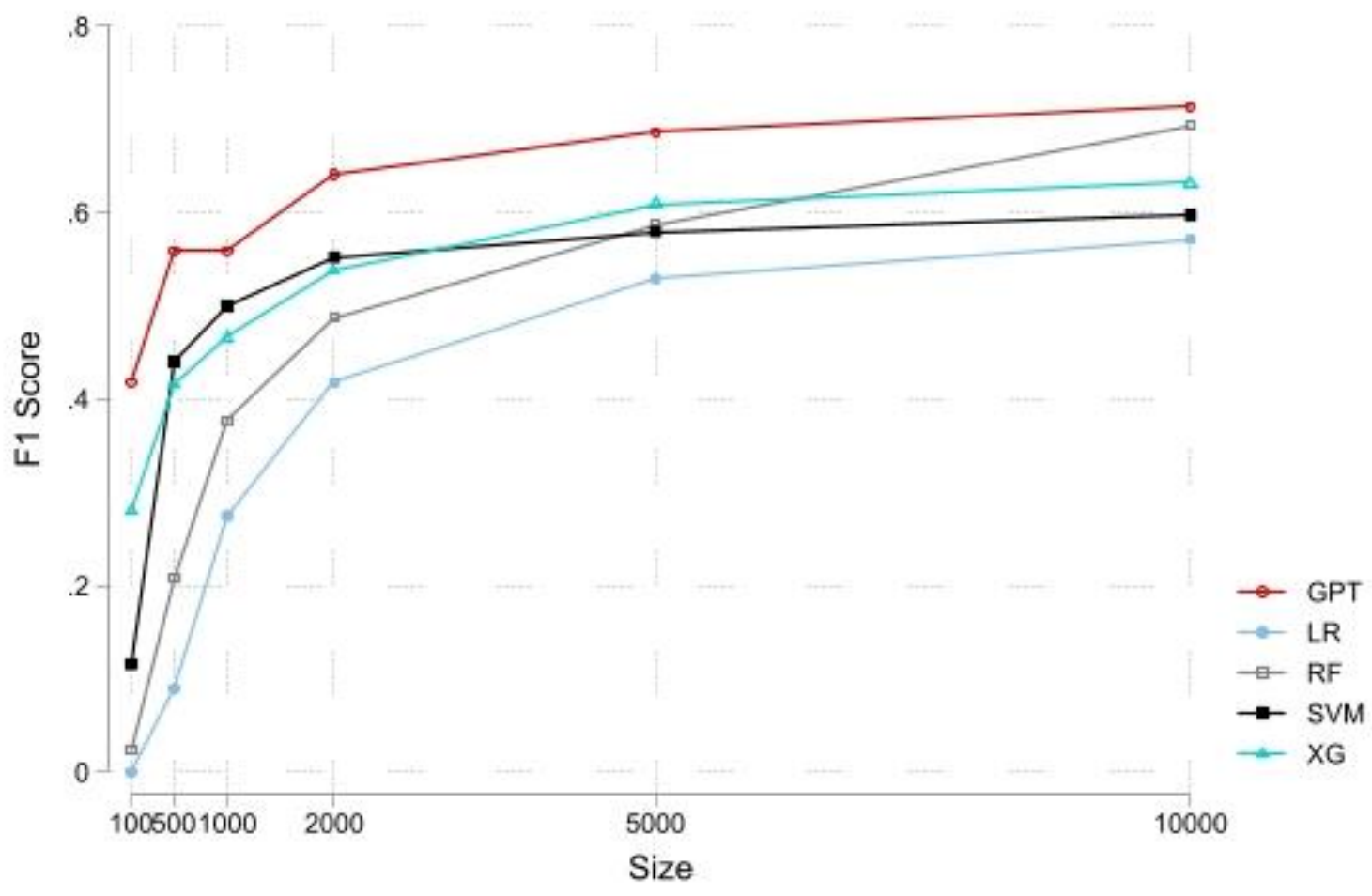

**Fig 2: F2 Scores from Full Dataset**

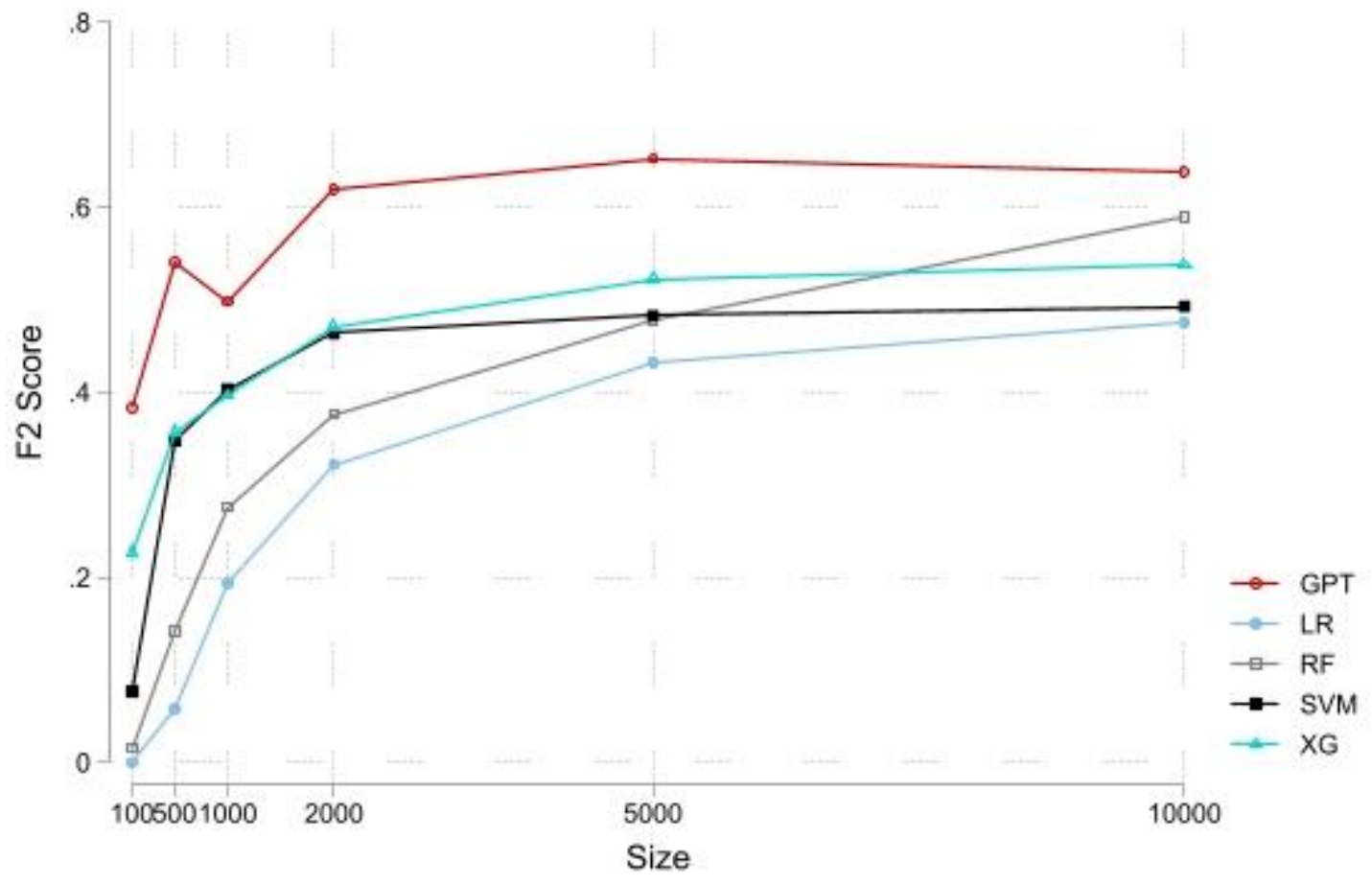

**Fig 3: F1 Scores from Cold Start Dataset**

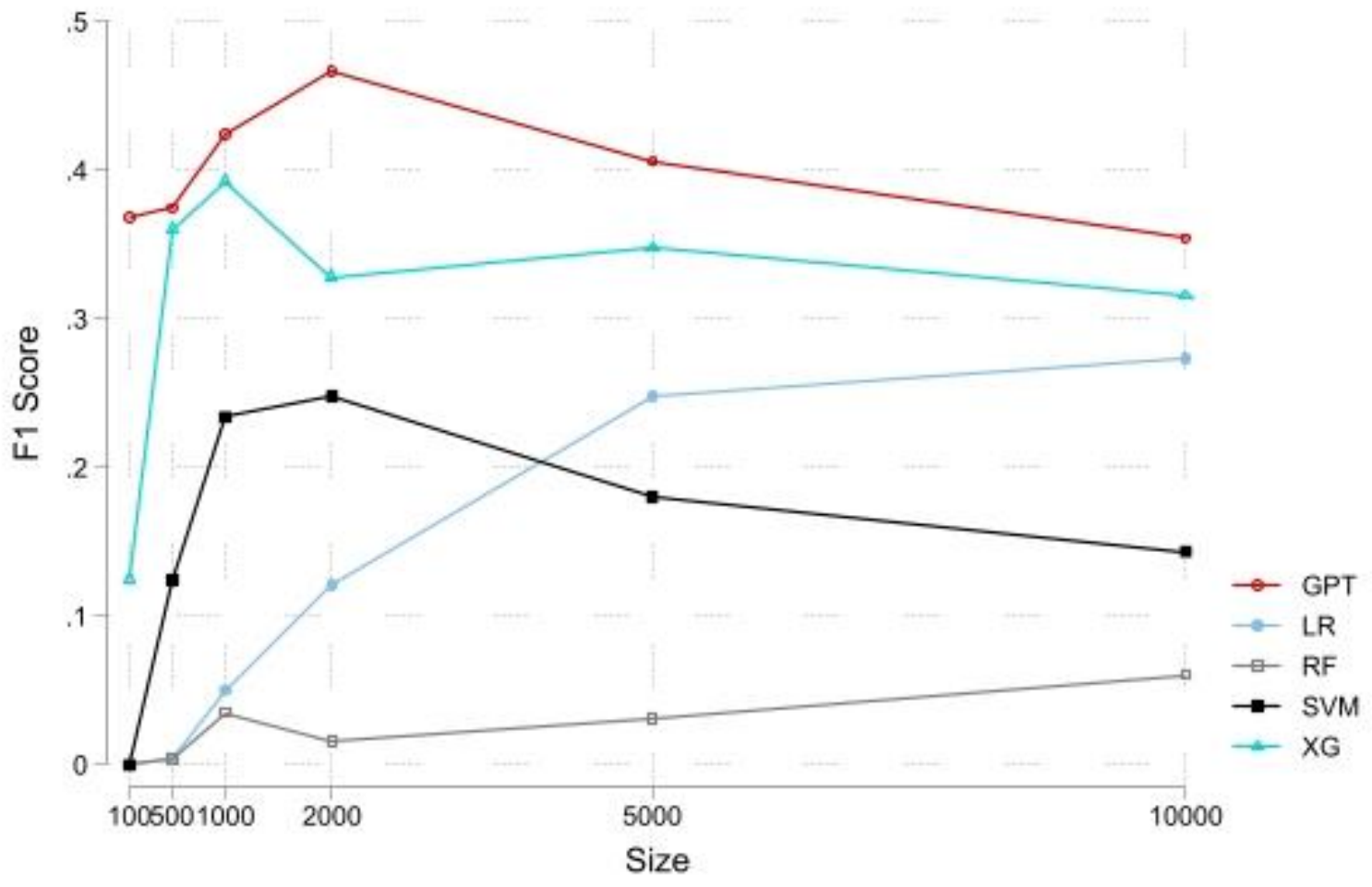

**Fig 4: F2 Scores from Cold Start Dataset**

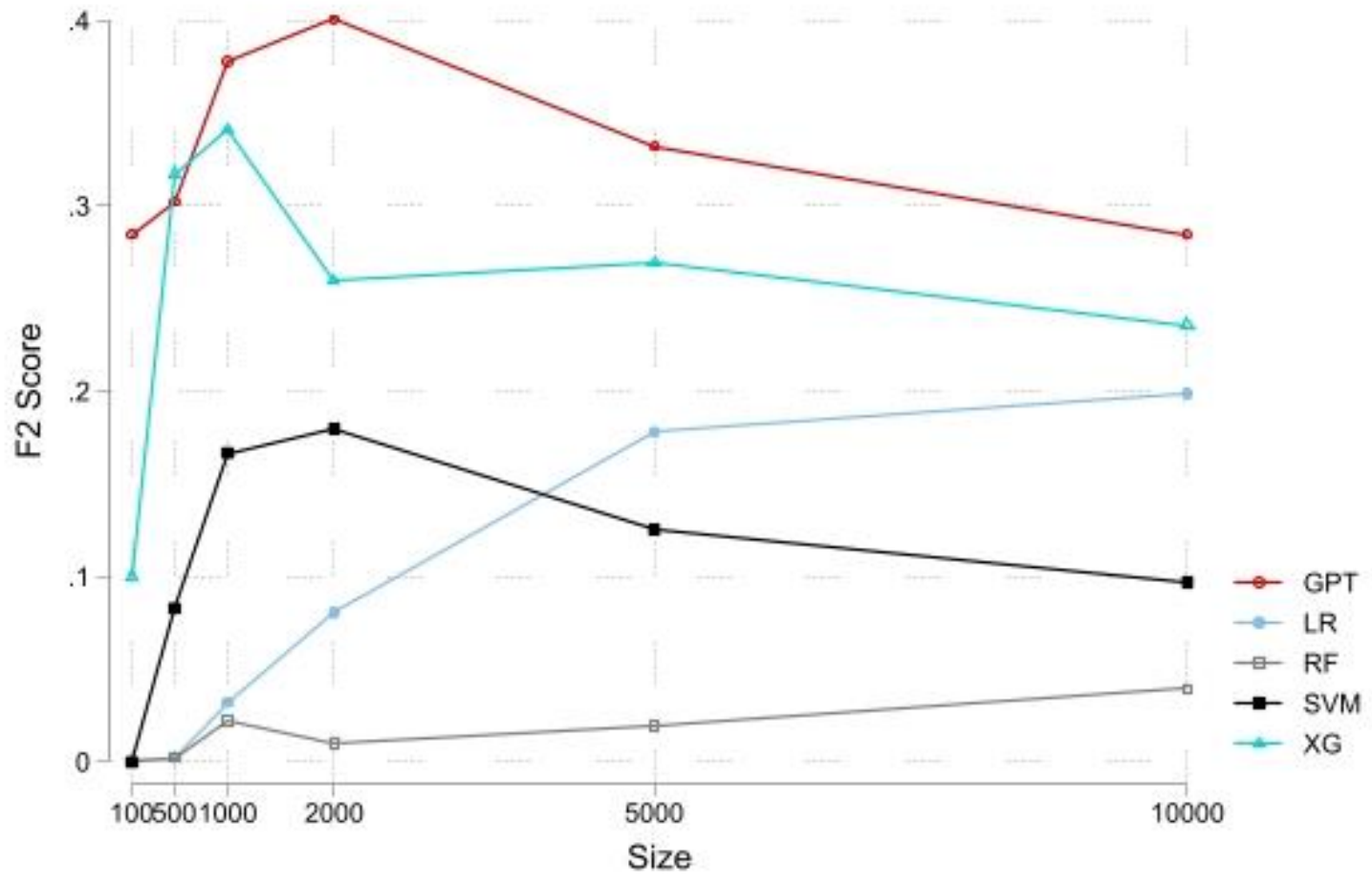

## REFERENCES

[1] B. Gu, J. Park, P. Konana, Research note—the impact of external word-of-mouth sources on retailer sales of high-involvement products, Inf. Syst. Res. 23 (2012) 182–196.

[2] M. Sun, How does the variance of product ratings matter?, Manag. Sci. 58 (2012) 696–707.

[3] L. Grewal, A.T. Stephen, In Mobile We Trust: The Effects of Mobile Versus Nonmobile Reviews on Consumer Purchase Intentions, J. Mark. Res. 56 (2019) 791–808. https://doi.org/10.1177/0022243719834514.

[4] R. Cui, J. Li, D.J. Zhang, Reducing Discrimination with Reviews in the Sharing Economy: Evidence from Field Experiments on Airbnb, Manag. Sci. 66 (2020) 1071–1094. https://doi.org/10.1287/mnsc.2018.3273.

[5] M.-X. Li, C.-H. Tan, K.-K. Wei, K.-L. Wang, Sequentiality of Product Review Information Provision: An Information Foraging Perspective, MIS Q. 41 (2017) 867–892.

[6] Y. Liu, J. Feng, X. Liao, When Online Reviews Meet Sales Volume Information: Is More or Accurate Information Always Better?, Inf. Syst. Res. 28 (2017) 723–743. https://doi.org/10.1287/isre.2017.0715.

[7] N. Hu, P.A. Pavlou, J.J. Zhang, On Self-Selection Biases in Online Product Reviews, MIS Q. 41 (2017) 449–471.

[8] N. Sahoo, C. Dellarocas, S. Srinivasan, The Impact of Online Product Reviews on Product Returns, Inf. Syst. Res. 29 (2018) 723–738. https://doi.org/10.1287/isre.2017.0736.

[9] K. Kim, K. Chung, N. Lim, Third-Party Reviews and Quality Provision, Manag. Sci. 65 (2019) 2695–2716. https://doi.org/10.1287/mnsc.2018.3082.

[10] C. Yi, Z. Jiang, X. Li, X. Lu, Leveraging User-Generated Content for Product Promotion: The Effects of Firm-Highlighted Reviews, Inf. Syst. Res. 30 (2019) 711–725.

[11] A.A. Choi, D. Cho, D. Yim, J.Y. Moon, W. Oh, When Seeing Helps Believing: The Interactive Effects of Previews and Reviews on E-Book Purchases, Inf. Syst. Res. 30 (2019) 1164–1183.

[12] D.D. Gunawan, K.-H. Huarng, Viral effects of social network and media on consumers' purchase intention, J. Bus. Res. 68 (2015) 2237–2241. https://doi.org/10.1016/j.jbusres.2015.06.004.

[13] E. Bigne, K. Chatzipanagiotou, C. Ruiz, Pictorial content, sequence of conflicting online reviews and consumer decision-making: The stimulus-organism-response model revisited, J. Bus. Res. 115 (2020) 403–416. https://doi.org/10.1016/j.jbusres.2019.11.031.

[14] A.D. Shukla, G. Gao, R. Agarwal, How Digital Word-of-Mouth Affects Consumer Decision Making: Evidence from Doctor Appointment Booking, Manag. Sci. 67 (2021) 1546–1568. https://doi.org/10.1287/mnsc.2020.3604.

[15] W. Duan, B. Gu, A.B. Whinston, Do online reviews matter? — An empirical investigation of panel data, Decis. Support Syst. 45 (2008) 1007–1016. https://doi.org/10.1016/j.dss.2008.04.001.

[16] X. Li, C. Wu, F. Mai, The effect of online reviews on product sales: A joint sentiment-topic analysis, Inf. Manage. 56 (2019) 172–184. https://doi.org/10.1016/j.im.2018.04.007.

[17] WOMMA, Return on Word of Mouth Study, (2014). http://womma.org/wp-content/uploads/2015/09/STUDY-WOMMA-Return-on-WOM-Executive-Summary.pdf.

[18] D. Lee, K. Hosanagar, How Do Product Attributes and Reviews Moderate the Impact of Recommender Systems Through Purchase Stages?, Manag. Sci. (2020). https://doi.org/10.1287/mnsc.2019.3546.

[19] T.W. Gruen, T. Osmonbekov, A.J. Czaplewski, eWOM: The impact of customer-to-customer online know-how exchange on customer value and loyalty, J. Bus. Res. 59 (2006) 449–456. https://doi.org/10.1016/j.jbusres.2005.10.004.

[20] Q.B. Liu, E. Karahanna, The Dark Side of Reviews: The Swaying Effects of Online Product Reviews on Attribute Preference Construction, MIS Q. 41 (2017) 427–448.

[21] C. Park, T.M. Lee, Information direction, website reputation and eWOM effect: A moderating role of product type, J. Bus. Res. 62 (2009) 61–67. https://doi.org/10.1016/j.jbusres.2007.11.017.

[22] M. Fetscherin, The five types of brand hate: How they affect consumer behavior, J. Bus. Res. 101 (2019) 116–127. https://doi.org/10.1016/j.jbusres.2019.04.017.

[23] M. Luca, G. Zervas, Fake It Till You Make It: Reputation, Competition, and Yelp Review Fraud, Manag. Sci. 62 (2016) 3412–3427. https://doi.org/10.1287/mnsc.2015.2304.
[24] D. Smith, Amazon reviewers brought to book, The Guardian. (2004). https://www.theguardian.com/technology/2004/feb/15/books.booksnews.
[25] D. Segal, A Rave, a Pan, or Just a Fake?, N. Y. Times. (2011). https://www.nytimes.com/2011/05/22/your-money/22haggler.html.
[26] U.M. Ananthakrishnan, B. Li, M.D. Smith, A Tangled Web: Should Online Review Portals Display Fraudulent Reviews?, Inf. Syst. Res. 31 (2020) 950–971. https://doi.org/10.1287/isre.2020.0925.
[27] Attorney General Report, (2013). http://www.ag.ny.gov/press-release/ag-schneiderman-announces-agreement-19-companies-stop-writing-fake-online-reviews-and.
[28] FTC Puts Hundreds of Businesses on Notice about Fake Reviews and Other Misleading Endorsements, Fed. Trade Comm. (2021). https://www.ftc.gov/news-events/news/press-releases/2021/10/ftc-puts-hundreds-businesses-notice-about-fake-reviews-other-misleading-endorsements (accessed March 29, 2023).
[29] S. Maheshwari, When Is a Star Not Always a Star? When It's an Online Review., N. Y. Times. (n.d.). https://www.nytimes.com/2019/11/28/business/online-reviews-fake.html.
[30] E. Woollacott, Amazon's Fake Review Problem Is Getting Worse, Forbes. (2019). https://www.forbes.com/sites/emmawoollacott/2019/04/16/amazons-fake-review-problem-is-getting-worse/?sh=320938e3195f.
[31] A. Mukherjee, B. Liu, J. Wang, N. Glance, N. Jindal, Detecting group review spam, in: ACM, 1963240, 2011: pp. 93–94. https://doi.org/10.1145/1963192.1963240.
[32] A. Mukherjee, B. Liu, N. Glance, Spotting fake reviewer groups in consumer reviews, in: ACM, 2187863, 2012: pp. 191–200. https://doi.org/10.1145/2187836.2187863.
[33] E.-P. Lim, V.-A. Nguyen, N. Jindal, B. Liu, H.W. Lauw, Detecting product review spammers using rating behaviors, in: ACM, 1871557, 2010: pp. 939–948. https://doi.org/10.1145/1871437.1871557.
[34] A. Mukherjee, V. Venkataraman, B. Liu, N. Glance, Fake Review Detection: Classification and Analysis of Real and Pseudo Reviews, University of Illinois at Chicago, Chicago, 2013. https://pdfs.semanticscholar.org/4c52/1025566e6afceb9adcf27105cd33e4022fb6.pdf.
[35] M. Ott, Y. Choi, C. Cardie, J.T. Hancock, Finding deceptive opinion spam by any stretch of the imagination, in: Association for Computational Linguistics, 2002512, 2011: pp. 309–319.
[36] R. Mihalcea, C. Strapparava, The lie detector: explorations in the automatic recognition of deceptive language, in: Association for Computational Linguistics, 1667679, 2009: pp. 309–312.
[37] A. Mukherjee, V. Venkataraman, B. Liu, N. Glance, What yelp fake review filter might be doing?, Assoc. Adv. Artif. Intell. AAAI. (2013) 409–418.
[38] Q.V. Le, T. Mikolov, Distributed Representations of Sentences and Documents, ArXiv. (2014). https://doi.org/10.48550/ARXIV.1405.4053.
[39] D. Tang, B. Qin, T. Liu, Document Modeling with Gated Recurrent Neural Network for Sentiment Classification, in: 2015: pp. 1422–1432.
[40] X. Tang, T. Qian, Z. You, Generating behavior features for cold-start spam review detection with adversarial learning, Inf. Sci. 526 (2020) 274–288. https://doi.org/10.1016/j.ins.2020.03.063.
[41] X. Wang, K. Liu, J. Zhao, Handling Cold-Start Problem in Review Spam Detection by Jointly Embedding Texts and Behaviors, in: Proc. 55th Annu. Meet. Assoc. Comput. Linguist. Vol. 1 Long Pap., Association for Computational Linguistics, Vancouver, Canada, 2017: pp. 366–376. https://doi.org/10.18653/v1/P17-1034.
[42] L. Xiang, H. You, G. Guo, Q. Li, Deep feature fusion for cold-start spam review detection, J. Supercomput. 79 (2023) 419–434. https://doi.org/10.1007/s11227-022-04685-z.
[43] G.G. Gao, J.S. McCullough, R. Agarwal, A.K. Jha, A Changing Landscape of Physician Quality Reporting: Analysis of Patients' Online Ratings of Their Physicians Over a 5-Year Period, J. Med. Internet Res. 14 (2012) e38. https://doi.org/10.2196/jmir.2003.
[44] G. Gao, B.N. Greenwood, R. Agarwal, J.S. McCullough, Vocal Minority and Silent Majority: How Do Online Ratings Reflect Population Perceptions of Quality, MIS Q. 39 (2015) 565–590. https://doi.org/10.25300/Misq/2015/39.3.03.

[45] J. Daley, P.M. Gertman, T.L. Delbanco, Looking for Quality in Primary Care Physicians, Health Aff. (Millwood). 7 (1988) 107–13. https://doi.org/10.1377/hlthaff.7.1.107.
[46] C.J. Salisbury, How do people choose their doctor?, Br. Med. J. BMJ. 299 (1989). https://www.bmj.com/content/bmj/299/6699/608.full.pdf.
[47] D.W. Bates, A.A. Gawande, The Impact of The Internet On Quality Measurement, Health Aff. (Millwood). 19 (2000) 104–14. https://doi.org/10.1377/hlthaff.19.6.104.
[48] B.H. Bornstein, D. Marcus, W. Cassidy, Choosing a doctor: an exploratory study of factors influencing patients' choice of a primary care doctor, J. Eval. Clincal Pract. 6 (2001) 255–62. https://doi.org/10.1046/j.1365-2753.2000.00256.x.
[49] D.A. Hanauer, K. Zheng, D.C. Singer, Public Awareness, Perception, and Use of Online Physician Rating Sites, JAMA. 311 (2014) 734–5. https://doi.org/10.1001/jama.2013.283194.
[50] M. Emmert, F. Meier, F. Pisch, U. Sander, Physician Choice Making and Characteristics Associated With Using Physician-Rating Websites: Cross-Sectional Study, JMIR. 15 (2013). https://www.jmir.org/2013/8/e187/.
[51] A. Victoor, D.M. Delnoij, R.D. Friele, J.J. Rademakers, Determinants of patient choice of healthcare providers: a scoping review, BMC Health Serv. Res. 12 (2012) 272. https://doi.org/10.1186/1472-6963-12-272.
[52] D. King, S. Zaman, S.S. Zaman, G.K. Kahlon, A. Naik, A.S. Jessel, N. Nanavati, A. Shah, B. Cox, A. Darzi, Identifying Quality Indicators Used by Patients to Choose Secondary Health Care Providers: A Mixed Methods Approach, JMIR. 3 (2015). https://mhealth.jmir.org/2015/2/e65/.
[53] Y.T. Wun, T.P. Lam, K.F. Lam, D. Goldberg, D.K.T. Li, K.C. Yip, How do patients choose their doctors for primary care in a free market?, J. Eval. Clincal Pract. 16 (2010) 1215–20. https://doi.org/10.1111/j.1365-2753.2009.01297.x.
[54] S.M. Yassini, M.A. Harrazi, J. Askari, The study of most important factors influencing physician choice, Procedia - Scoial Behav. Sci. 5 (2010) 1945–1949. https://doi.org/10.1016/j.sbspro.2010.07.393.
[55] L. Hedges, How Patients Use Online Reviews, Gartner. (2019). https://www.softwareadvice.com/resources/how-patients-use-online-reviews/.
[56] C. Dellarocas, X.Q. Zhang, N.F. Awad, Exploring the value of online product reviews in forecasting sales: The case of motion pictures, J. Interact. Mark. 21 (2007) 23–45. https://doi.org/10.1002/dir.20087.
[57] W. Duan, B. Gu, A.B. Whinston, The dynamics of online word-of-mouth and product sales—An empirical investigation of the movie industry, J. Retail. 84 (2008) 233–242. https://doi.org/10.1016/j.jretai.2008.04.005.
[58] M. Trusov, R.E. Bucklin, K. Pauwels, Effects of Word-of-Mouth Versus Traditional Marketing: Findings from an Internet Social Networking Site, J. Mark. 73 (2009) 90–102. https://doi.org/10.1509/jmkg.73.5.90.
[59] P. Adamopoulos, A. Ghose, V. Todri, The Impact of User Personality Traits on Word of Mouth: Text-Mining Social Media Platforms, Inf. Syst. Res. 29 (2018) 612–640. https://doi.org/10.1287/isre.2017.0768.
[60] N. Hu, L. Liu, J. Zhang, Do online reviews affect product sales? The role of reviewer characteristics and temporal effects, Inf. Technol. Manag. 9 (2008) 201–214. https://doi.org/10.1007/s10799-008-0041-2.
[61] E.K. Clemons, G.G. Gao, L.M. Hitt, When online reviews meet hyperdifferentiation: A study of the craft beer industry, J. Manag. Inf. Syst. 23 (2006) 149–171. https://doi.org/10.2753/Mis0742-1222230207.
[62] D.K. Gauri, A. Bhatnagar, R. Rao, Role of word of mouth in online store loyalty, Commun. ACM. 51 (2008) 89–91. https://doi.org/10.1145/1325555.1325572.
[63] X. Luo, Quantifying the Long-Term Impact of Negative Word of Mouth on Cash Flows and Stock Prices, Mark. Sci. 28 (2009) 148–165. https://doi.org/10.1287/mksc.1080.0389.
[64] Mintel Research, (2015). http://www.mintel.com/blog/mintel-market-news/mintel-in-the-media-this-weeks-highlights-23-october-2015.
[65] N. McCarthy, Americans Visit Their Doctor 4 Times A Year. People In Japan Visit 13 Times A Year, Forbes. (2014). https://www.forbes.com/sites/niallmccarthy/2014/09/04/americans-visit-their-doctor-4-times-a-year-people-in-japan-visit-13-times-a-year-infographic/#e43eaa6e3475.
[66] W. Emons, Credence Goods and Fraudulent Experts, RAND J. Econ. 28 (1997) 107–119. https://doi.org/10.2307/2555942.

[67] J.M. Austin, A.K. Jha, P.S. Romano, S.J. Singer, T.J. Vogus, R.M. Wachter, P.J. Pronovost, National hospital ratings systems share few common scores and may generate confusion instead of clarity, Health Aff Millwood. 34 (2015) 423–30. https://doi.org/10.1377/hlthaff.2014.0201.
[68] S. Saghafian, W.J. Hopp, Can Public Reporting Cure Healthcare? The Role of Quality Transparency in Improving Patient–Provider Alignment, Oper. Res. 68 (2020) 71–92. https://doi.org/10.1287/opre.2019.1868.
[69] M. Faber, M. Bosch, H. Wollersheim, S. Leatherman, R. Grol, Public reporting in health care: how do consumers use quality-of-care information? A systematic review, Med Care. 47 (2009) 1–8. https://doi.org/10.1097/MLR.0b013e3181808bb5.
[70] Medical Council of India, Professional Conduct, Etiquette and Ethics Regulations, (2002).
[71] N. Jindal, B. Liu, Analyzing and detecting review spam, in: IEEE, 2007: pp. 547–552.
[72] N. Jindal, B. Liu, Opinion spam and analysis, in: ACM, 1341560, 2008: pp. 219–230. https://doi.org/10.1145/1341531.1341560.
[73] G. Wang, S. Xie, B. Liu, P.S. Yu, Review Graph Based Online Store Review Spammer Detection, in: Vancouver, Canada, 2011: pp. 1242–1247. https://doi.org/10.1109/ICDM.2011.124.
[74] B. Liu, Sentiment analysis and opinion mining, Synth. Lect. Hum. Lang. Technol. 5 (2012) 1–167.
[75] N. Jindal, B. Liu, Review spam detection, in: ACM, 1242759, 2007: pp. 1189–1190. https://doi.org/10.1145/1242572.1242759.
[76] O. for C. Rights (OCR), Summary of the HIPAA Privacy Rule, HHS.Gov. (2008). https://www.hhs.gov/hipaa/for-professionals/privacy/laws-regulations/index.html (accessed March 18, 2023).
[77] M.K. Johnson, C.L. Raye, Reality monitoring, Psychol. Rev. 88 (1981) 67–85. https://doi.org/10.1037/0033-295X.88.1.67.
[78] A. Vrij, S. Mann, S. Kristen, R.P. Fisher, Cues to deception and ability to detect lies as a function of police interview styles, Law Hum. Behav. 31 (2007) 499–518.
[79] S. Xie, G. Wang, S. Lin, P.S. Yu, Review spam detection via temporal pattern discovery, in: ACM, 2339662, 2012: pp. 823–831. https://doi.org/10.1145/2339530.2339662.
[80] W. Jabr, The Fallacy of Spam Reviews, in: Workshop Inf. Syst. Econ., University of Texas at Dallas., 2015.
[81] S. Feng, R. Banerjee, Y. Choi, Syntactic stylometry for deception detection, in: Association for Computational Linguistics, 2390708, 2012: pp. 171–175.
[82] J. Li Ott, M.,. Cardie, C.,.&. Hovy, E.H., Towards a General Rule for Identifying Deceptive Opinion Spam., in: 2014.
[83] A. Neelakantan, T. Xu, R. Puri, A. Radford, J.M. Han, J. Tworek, Q. Yuan, N. Tezak, J.W. Kim, C. Hallacy, J. Heidecke, P. Shyam, B. Power, T.E. Nekoul, G. Sastry, G. Krueger, D. Schnurr, F.P. Such, K. Hsu, M. Thompson, T. Khan, T. Sherbakov, J. Jang, P. Welinder, L. Weng, Text and Code Embeddings by Contrastive Pre-Training, (2022). https://doi.org/10.48550/arXiv.2201.10005.
[84] Z. Bu, Z. Xia, J. Wang, A sock puppet detection algorithm on virtual spaces, Knowl.-Based Syst. 37 (2013) 366–377. https://doi.org/10.1016/j.knosys.2012.08.016.
[85] A. Flood, Sock puppetry and fake reviews: publish and be damned, The Guardian. (2012). https://www.theguardian.com/books/2012/sep/04/sock-puppetry-publish-be-damned (accessed March 20, 2023).
[86] A. Radford, K. Narasimhan, T. Salimans, I. Sutskever, Improving language understanding by generative pre-training, (2018).
[87] E. Gilbert, K. Karahalios, Understanding deja reviewers, in: ACM, 1718961, 2010: pp. 225–228. https://doi.org/10.1145/1718918.1718961.
[88] T. Brown, B. Mann, N. Ryder, M. Subbiah, J.D. Kaplan, P. Dhariwal, A. Neelakantan, P. Shyam, G. Sastry, A. Askell, S. Agarwal, A. Herbert-Voss, G. Krueger, T. Henighan, R. Child, A. Ramesh, D. Ziegler, J. Wu, C. Winter, C. Hesse, M. Chen, E. Sigler, M. Litwin, S. Gray, B. Chess, J. Clark, C. Berner, S. McCandlish, A. Radford, I. Sutskever, D. Amodei, Language Models are Few-Shot Learners, in: Adv. Neural Inf. Process. Syst., Curran Associates, Inc., 2020: pp. 1877–1901.

https://proceedings.neurips.cc/paper/2020/hash/1457c0d6bfcb4967418bfb8ac142f64a-Abstract.html (accessed March 13, 2023).
[89] P. Craja, A. Kim, S. Lessmann, Deep learning for detecting financial statement fraud, Decis. Support Syst. 139 (2020) 113421. https://doi.org/10.1016/j.dss.2020.113421.
[90] Q. Li, H. Peng, J. Li, C. Xia, R. Yang, L. Sun, P.S. Yu, L. He, A Survey on Text Classification: From Traditional to Deep Learning, ACM Trans. Intell. Syst. Technol. 13 (2022) 31:1-31:41. https://doi.org/10.1145/3495162.
[91] T. Joachims, A support vector method for multivariate performance measures, in: Proc. 22nd Int. Conf. Mach. Learn., 2005: pp. 377–384.
[92] S. Banerjee, A.Y.K. Chua, J.-J. Kim, Don't be deceived: Using linguistic analysis to learn how to discern online review authenticity, J. Assoc. Inf. Sci. Technol. 68 (2017) 1525–1538. https://doi.org/10.1002/asi.23784.
[93] L. Zhou, J.K. Burgoon, D.P. Twitchell, T. Qin, J.F. NUNAMAKER Jr., A Comparison of Classification Methods for Predicting Deception in Computer-Mediated Communication, J. Manag. Inf. Syst. 20 (2004) 139–166. https://doi.org/10.1080/07421222.2004.11045779.
[94] D. Mayzlin, Y. Dover, J. Chevalier, Promotional Reviews: An Empirical Investigation of Online Review Manipulation, Am. Econ. Rev. 104 (//) 2421–2455. https://doi.org/10.1257/aer.104.8.2421.
[95] G. Fei, A. Mukherjee, B. Liu, M. Hsu, M. Castellanos, R. Ghosh, Exploiting Burstiness in Reviews for Review Spammer Detection, Proc. Int. AAAI Conf. Web Soc. Media. 7 (2021) 175–184. https://doi.org/10.1609/icwsm.v7i1.14400.
[96] R. Barbado, O. Araque, C.A. Iglesias, A framework for fake review detection in online consumer electronics retailers, Inf. Process. Manag. 56 (2019) 1234–1244. https://doi.org/10.1016/j.ipm.2019.03.002.
[97] C. Sun, Q. Du, G. Tian, Exploiting Product Related Review Features for Fake Review Detection, Math. Probl. Eng. 2016 (2016) 1–7. https://doi.org/10.1155/2016/4935792.
[98] J. Fontanarava, G. Pasi, M. Viviani, Feature Analysis for Fake Review Detection through Supervised Classification, in: 2017 IEEE Int. Conf. Data Sci. Adv. Anal. DSAA, IEEE, Tokyo, Japan, 2017: pp. 658–666. https://doi.org/10.1109/DSAA.2017.51.
[99] L. Akoglu, R. Chandy, C. Faloutsos, Opinion Fraud Detection in Online Reviews by Network Effects, in: 2013: pp. 2–11.
[100] K.A. Appiah, Can Therapists Fake Their Own Online Reviews?, N. Y. Times Mag. (2017). https://www.nytimes.com/2017/02/22/magazine/can-therapists-fake-their-own-online-reviews.html.
[101] J.C. Wong, Anti-vaxx 'mobs': doctors face harassment campaigns on Facebook, The Guardian. (2019). https://www.theguardian.com/technology/2019/feb/27/facebook-anti-vaxx-harassment-campaigns-doctors-fight-back.
[102] H. Bodkin, GPs are posing as patients and posting 'fake reviews' online, health chiefs reveal, The Telegraph. (2018). https://www.telegraph.co.uk/news/2018/07/16/gps-posting-fake-reviews-online-health-chiefs-reveal/.
[103] J. Fleischer, R. Yarborough, J. Piper, Doctors: Get a Second Opinion Before Trusting Reviews, NBC Wash. (2018). https://www.nbcwashington.com/investigations/Doctors-Get-a-Second-Opinion-Before-Trusting-Online-Medical-Reviews-498936581.html.
[104] Fake patient reviews are making it increasingly hard to seek medical help on Google, Yelp and other directory sites, Wash. Post. (2021). https://www.washingtonpost.com/business/2021/06/04/fake-medical-reviews-google-zocdoc-trustpilot/ (accessed April 1, 2023).
[105] Choosing a new doctor? Beware of fake rave reviews, TODAY.Com. (2022). https://www.today.com/health/health/fake-doctor-reviews-rcna39193 (accessed April 1, 2023).
[106] Doctors Using Fake Positive Reviews to Boost Business, Medscape. (2022). https://www.medscape.com/viewarticle/979310 (accessed April 1, 2023).